\documentclass[conference]{IEEEtran}
\IEEEoverridecommandlockouts

\usepackage[
    backend=biber,
    style=ieee,
    sorting=none,
    natbib=true,
    doi=false,
    isbn=false,
    url=false,
    eprint=false,
    maxcitenames=2,
    mincitenames=1,
    citestyle=numeric-comp
]{biblatex}
\usepackage[pdftex,colorlinks]{hyperref}

\usepackage[printonlyused]{acronym}

\usepackage{siunitx}
\usepackage[all]{nowidow}

\usepackage{tikz}
\usetikzlibrary{arrows}
\usetikzlibrary{arrows.meta}
\usepackage{makecell}
\usetikzlibrary{positioning}
\usetikzlibrary{decorations.pathreplacing,calligraphy}
\usepackage[normalem]{ulem}

\usepackage{lipsum}

\usepackage{hhline}

\usepackage{xspace} 

\usepackage{epstopdf}

\usepackage{import}

\usepackage[table]{xcolor}
\usepackage{booktabs}
\usepackage{array}
\usepackage{longtable}
\newcolumntype{L}{>{\raggedright\arraybackslash}p{0.13\textwidth}}
\newcolumntype{C}{>{\centering\arraybackslash}p{0.10\textwidth}}
\newcolumntype{Y}{>{\centering\arraybackslash}m{1.8cm}}
\newcolumntype{M}{>{\centering\arraybackslash}p{0.065\textwidth}}

\usepackage{tabularx}
\usepackage{multirow, multicol}

\newlength{\Oldarrayrulewidth}

\newcolumntype{?}[1]{!{\vrule width #1}}

\usepackage{amssymb,amsfonts,amsmath,amscd}

\usepackage{bm}

\usepackage[T1]{fontenc}

\usepackage{cancel}

\newcommand{\bbm}{\begin{bmatrix}}
\newcommand{\ebm}{\end{bmatrix}}

\newcommand{\ignore}[1]{}

\newcommand{\bma}[1]{\left[\begin{array}{#1}}
\newcommand{\ema}{\end{array}\right]}

\DeclareMathAlphabet{\mbf}{OT1}{ptm}{b}{n}

\def\fdotb{{\raisebox{-0.6ex}{ \kern0.2ex\raisebox{0.8ex}{\tiny $\hspace*{-1ex}\circ$}}}}
\def\fddotb{{\raisebox{-0.6ex}{ \kern0.2ex\raisebox{0.8ex}{\tiny $\hspace*{-1ex}\circ\circ$}}}}

\newcommand{\trans}{{\ensuremath{\mathsf{T}}}} 
\newcommand{\utimes}{ {\raisebox{-0.6ex}{ \kern-1.0ex\raisebox{0.6ex}{ \small $\mathsf{v}$}}} } %
\newcommand{\beq}{\begin{equation}}
\newcommand{\eeq}{\end{equation}}
\newcommand{\bdis}{\begin{displaymath}}
\newcommand{\edis}{\end{displaymath}}
\newcommand{\beqarray}{\begin{eqnarray}}
\newcommand{\eeqarray}{\end{eqnarray}}
\newcommand{\beqarraynn}{\begin{eqnarray*}}
\newcommand{\eeqarraynn}{\end{eqnarray*}}

\DeclareMathAlphabet{\mbf}{OT1}{ptm}{b}{n}

\newcommand{\maev}{\mathrm{MAE}_{V}}

\DeclareMathSizes{6.5}{6.5}{5}{5}

\begin{document}

\title{Do Spinning Radar Doppler Velocity Measurements Improve Vehicle Detection and Tracking?\\
\thanks{$^1$ Robotics Institute,
University of Toronto, Canada \\
Corresponding author: Eric Xie, ericx.xie@mail.utoronto.ca}
}


\author{\IEEEauthorblockN{Eric Xie$^{1}$, Daniil Lisus$^{1}$, Timothy D. Barfoot$^{1}$}
}

\maketitle

\begin{abstract}
Spinning frequency-modulated continuous-wave (FMCW) radars have been gaining popularity in autonomous vehicle perception on account of their robustness to adverse weather conditions and $360^{\circ}$ field of view.
Recently, scanning radars have also been shown capable of generating per-azimuth Doppler velocity.
In this paper, we investigate whether these Doppler velocity measurements improve spinning radar vehicle detection and tracking performance.
For detection, we estimate the ego motion and use it to undo the Doppler range distortion of the radar image before passing it to a network.
For tracking, we propose a new way to estimate a per-vehicle velocity and use it as a prior for the tracker's motion model.
Since Doppler-enabled spinning radar data is not available in any dataset with ground-truth dynamic object labels, our first contribution is an automatic labelling pipeline that uses an ensemble of fine-tuned off-the-shelf lidar detectors to label all \SI{643}{\kilo\metre} of the Boreas Road Trip dataset.
We then transfer detections to radar, and use over \SI{250}{\kilo\metre} of vehicle-dense sequences as ground-truth training data.
By training and evaluating two state-of-the-art detectors, we show that Doppler undistortion can improve detection accuracy by up to $2.37$ points on mean average precision.
Furthermore, we show that the Doppler velocity prior can improve tracking accuracy by $13.68$ points on multi-object tracking accuracy (MOTA) versus the zero-velocity initialization baseline, while achieving $99.7\%$ of the MOTA obtained using ground-truth velocities as the prior. \textit{Preprint, under review.}

\end{abstract}


\section{Introduction}
\label{sec:intro}

Autonomous vehicles rely on range sensors to detect and track other vehicles around them. 
Lidar is a popular choice, because it is fast, accurate, and supported by a mature ecosystem of detectors and benchmarks~\cite{caesar_nuscenes_2020, sun_scalability_2020}. 
However, lidar is an optical sensor, so rain, snow, fog, smoke, and general obstacles can scatter its beams, making it unreliable in adverse weather conditions and scenarios with occlusion. 
Spinning frequency-modulated continuous-wave (FMCW) radar complements lidar, because its millimetre-wave signal can propagate around obstacles, return measurements at ranges longer than lidar, and resolve vehicles that are occluded to optical sensors~\cite{harlow_new_2024}. 
Its unique advantage allows it to make detections and sustain tracks that are otherwise lost to lidar. 
Furthermore, spinning radar provides dense, $360^{\circ}$ imagery from a single sensor, avoiding the inter-sensor calibration needed for coverage with multiple 4D radars.
Despite these advantages, three problems remain in spinning radar detection and tracking: learning supervision, Doppler distortion, and inter-frame motion.

\begin{figure}[!t]
  \centering
  \includegraphics[width=\columnwidth]{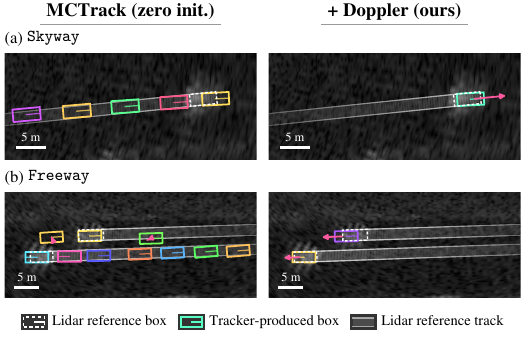}
  \caption{Tracking with the baseline's zero-velocity initialization (left) and with our Doppler velocity prior (right) on two Boreas-RT sequences. Box colour denotes track identity, and magenta arrows show the estimated velocity. The reference car in (a) moves to the right at \SI{35}{\metre/\second}, and the two in (b) to the left at \SI{25.9}{\metre/\second} and \SI{29.8}{\metre/\second}, respectively. The baseline tracker is unable to associate the radar labels across frames, causing track fragmentation. With our Doppler prior, the detections are associated into single continuous tracks.}
  \label{fig:qual-tracking}
\end{figure}

The first problem is learning supervision. 
The RADIATE dataset~\cite{sheeny_radiate_2020}, a popular dataset used for developing spinning radar detection methods, involves manual labelling of radar images. 
Manual annotation is slow, expensive, and hard to scale, limiting RADIATE to three hours of driving, and it can introduce frame-to-frame jitter inconsistent with a vehicle's true trajectory. 
Meanwhile, offboard auto-labelling has matured in the lidar world, where detection ensembles and offline trackers now produce labels competitive with human annotators~\cite{qi_offboard_2021,ma_detzero_2023,tsai_ms3d_2023}. 
However, comparable auto-labelling detection pipelines remain undeveloped for spinning radar.
We bridge this gap by creating a fully automatic pipeline built on the Boreas Road Trip (Boreas-RT) dataset~\cite{lisus_boreas_2026}, which consists of \SI{643}{\kilo\metre} of driving with a lidar and a Doppler-enabled spinning radar. 
Our pipeline combines lidar detection ensembles, offline tracking, and continuous-time trajectory refinement to generate temporally consistent radar labels.
We use these labels to train and evaluate two state-of-the-art spinning radar detectors, RaFD~\cite{yang_rafd_2025} and SIRA~\cite{yataka_sira_2024}, on Boreas-RT.

The second problem is Doppler distortion. 
In FMCW radar, radial motion relative to the sensor shifts returns away from their true ranges~\cite{burnett_we_2021}.
Labels drawn directly on radar returns can inherit their Doppler-induced range shifts, so evaluation against these labels may not reveal errors in physical position.
This becomes a problem for cross-modal supervision, because these shifts can misalign radar returns with boxes transferred from lidar.
We explicitly address this mismatch by estimating ego velocity from each scan's Doppler measurements and correcting the ego-induced range shifts before detection.

The third problem is inter-frame motion. 
A spinning radar typically completes a scan at \SI{4}{\hertz}, so a vehicle travelling at highway speed can move more than a car length between consecutive scans. 
Trackers that infer velocity by differencing poses, as commonly done in tracking-by-detection frameworks~\cite{weng_3d_2020,pang_simpletrack_2023,wang_mctrack_2024}, frequently fail at such speeds, resulting in fragmented tracks.
One solution is to leverage Doppler measurements, which provide velocity information directly, aiding inter-frame motion recovery.
While Doppler-enabled spinning radar has been studied for odometry~\cite{lisus_are_2024,legentil_dro_2025}, its use for object tracking remains unexplored. 
We address this gap with a novel method that derives per-vehicle velocity from Doppler measurements for track initialization.
This prior improves tracking over the baseline zero-velocity initialization (zero init.), particularly for fast-moving vehicles, and approaches the performance of a privileged ground-truth velocity prior.

The three problems motivate each of our three contributions:
\begin{enumerate}[\setlength{\itemsep}{0pt}\setlength{\parsep}{0pt}\setlength{\topsep}{2pt}]
\item For learning supervision, we introduce an automatic lidar-to-radar labelling pipeline to produce temporally consistent labels for spinning radar detection.

\item For Doppler distortion, we show that using the estimated ego velocity for Doppler correction of the spinning radar images improves detection.

\item For inter-frame motion, we propose a Doppler-based method for extracting per-vehicle velocity from a single scan and show that it improves tracking.
\end{enumerate}

\section{Related Work}
\label{sec:related}

\subsection{Offboard Auto-Labelling with Lidar}
\label{sec:related-labels}

\citet{qi_offboard_2021} establish offboard auto-labelling with lidar, where detector labels are tracked and refined offline using full temporal context, achieving near-human quality.
DetZero~\cite{ma_detzero_2023}, CTRL~\cite{fan_once_2023}, and LabelFormer~\cite{yang_labelformer_2023} follow the same detect, track, and refine structure.
While these pipelines assume the detector was trained on human labels from the target domain, MS3D++~\cite{tsai_ms3d_2023} tackles cross-domain adaptation by fusing an ensemble of detectors pretrained on other datasets and refining the fused boxes temporally.
Since Boreas-RT is out of distribution relative to the lidar detectors' pretraining data, we adopt the MS3D++ labelling framework because it targets this exact problem.
For tracking, these pipelines typically use a tracking-by-detection Kalman filter tracker such as AB3DMOT~\cite{weng_3d_2020}, SimpleTrack~\cite{pang_simpletrack_2023}, or MCTrack~\cite{wang_mctrack_2024}, the last of which we adopt for its strong benchmark performance.

\subsection{Lidar-to-Radar Label Transfer}
\label{sec:related-transfer}

Lidar has also been used to label radar directly.
Automotive radar detectors have been trained on boxes generated from lidar, semi-automatically~\cite{major_vehicle_2019}, or automatically~\cite{sun_efficient_2024}.
These works share our goal, but not our timing problem.
Automotive radars capture a frame in one short burst, so every azimuth effectively shares one timestamp.
They also operate at a higher frame rate on account of a narrower field of view, keeping the nearest-frame transfer error minimal.
For spinning radar, apart from its lower data frequency, each azimuth is captured at a different time, so frame-to-frame transfer can misplace a moving vehicle by metres.
MVDNet~\cite{qian_mvdnet_2021} addresses this problem on the Oxford Radar RobotCar dataset~\cite{barnes_oxford_2020} by keeping only the sector of each lidar scan swept at the same time as the radar, which can discard valid data.
Its boxes are also manually labelled every twentieth lidar frame and interpolated between annotations, which may introduce error from non-linear motion.
We instead fit each track as a continuous-time $\mathrm{SE}(3)$ trajectory~\cite{barfoot_batch_2014,anderson_full_2015} and sample it at the time the radar beam crosses each vehicle, which accounts for vehicle motion between lidar and radar observations.

\begin{figure*}[!t]
  \centering
\begingroup%
\def\labellingassetdir{tikz_images/labelling_figure}%
\def\labellingradarbadgeanchor{north west}%
\def\labellingradarbadgex{14.24}%
\definecolor{lpink}{HTML}{000000}%
\definecolor{lpgrey}{HTML}{627184}%
\definecolor{lplight}{HTML}{E4E9EF}%
\definecolor{lpcyan}{HTML}{59DFD0}%
\definecolor{lpfused}{HTML}{AEFFF6}%
\begin{tikzpicture}[
  font=\fontsize{7}{8.5}\selectfont,
  >={Stealth[length=2mm,width=1.6mm]},
  heading/.style={anchor=south west,inner sep=0pt,text=lpink,
    text height=1.8ex,text depth=.4ex,
    font=\fontsize{8}{9.5}\selectfont\bfseries},
  flow/.style={->,draw=lpgrey,line width=.9pt},
]
\foreach \x/\file in {0/01_lidar,3.56/02_ensemble,7.12/03_tracking,10.68/04_refinement,14.24/05_radar}{
  \node[anchor=south west,inner sep=0pt] at (\x,0)
    {\includegraphics[width=3.25cm,height=3.25cm]{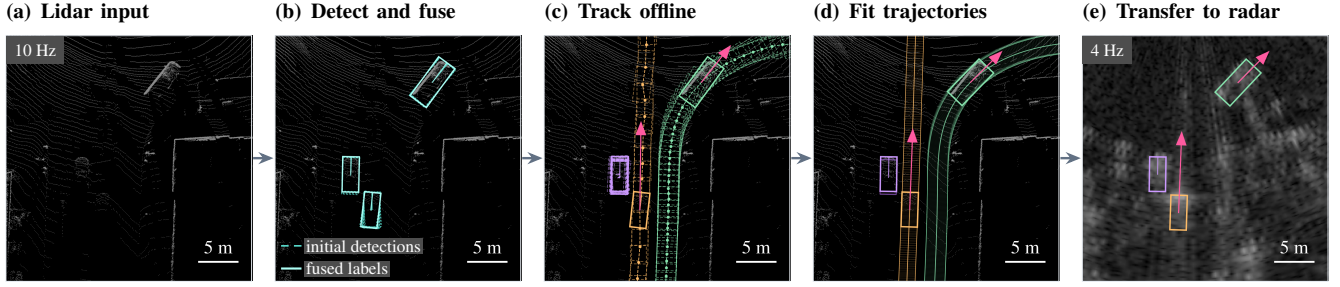}};
  \draw[draw=lplight,line width=.35pt] (\x,0) rectangle ++(3.25,3.25);
}
\node[heading] at (0,3.41) {(a)\hspace{.45em}Lidar input};
\node[heading] at (3.56,3.41) {(b)\hspace{.45em}Detect and fuse};
\node[heading] at (7.12,3.41) {(c)\hspace{.45em}Track offline};
\node[heading] at (10.68,3.41) {(d)\hspace{.45em}Fit trajectories};
\node[heading] at (14.24,3.41) {(e)\hspace{.45em}Transfer to radar};

\foreach \x in {3.27,6.83,10.39,13.95}{
  \draw[flow] (\x,1.63) -- ++(.27,0);
}
\draw[lpcyan,line width=.6pt,line cap=butt] (3.64,.47) -- (3.74,.47);
\draw[lpcyan,line width=.6pt,line cap=butt] (3.81,.47) -- (3.91,.47);
\draw[lpfused,line width=.8pt] (3.64,.17) -- (3.91,.17);
\node[anchor=west,inner sep=.7pt,text=white,fill=black!70,font=\fontsize{6.5}{7.5}\selectfont]
  at (3.94,.47) {initial detections};
\node[anchor=west,inner sep=.7pt,text=white,fill=black!70,font=\fontsize{6.5}{7.5}\selectfont]
  at (3.94,.17) {fused labels};
\node[anchor=north west,inner sep=3pt,text=white,fill=black!70,
  font=\fontsize{6.5}{7.5}\selectfont] at (0,3.25) {10 Hz};
\node[anchor=\labellingradarbadgeanchor,inner sep=3pt,text=white,fill=black!70,
  font=\fontsize{6.5}{7.5}\selectfont] at (\labellingradarbadgex,3.25) {4 Hz};
\end{tikzpicture}%
\endgroup%
  \caption{Automatic lidar-to-radar labelling on an \texttt{Urban} sequence of Boreas-RT.
  Starting from the \SI{10}{\hertz} lidar input (a), stage (b) detects and fuses proposals (cyan, dashed then solid) for three vehicles with ticks indicating box heading. Stage (c) generates tracks, where purple, orange, and green identify a stationary, straight-moving, and turning vehicle, respectively. Velocity arrows span the distance travelled in one second. Stage (d) refines tracks with a continuous-time $\mathrm{SE}(3)$ trajectory, and stage (e) samples the trajectory at the radar azimuth timestamp and transfers the resulting boxes to the \SI{4}{\hertz} radar scans.}
  \label{fig:labelling}
\end{figure*}

\subsection{Object Detection on Spinning Radar}
\label{sec:related-detection}

Object detection with spinning FMCW radar predates deep learning, with \citet{vivet_mobile_2012} comparing up-chirp and down-chirp spectra to detect moving objects.
Learned detectors instead require labelled scans, and RADIATE~\cite{sheeny_radiate_2020} remains the only dataset with scanning radar images labelled at scale, so the detectors closest to our work were all developed on it.
TempoRadar~\cite{li_exploiting_2022} introduces a temporal relational layer that associates object features across two consecutive radar frames.
SCTR~\cite{yataka_radar_2024} extends the relation to several frames, while SIRA~\cite{yataka_sira_2024} adds a motion-consistency track for association.
RaFD~\cite{yang_rafd_2025} estimates a bird's-eye-view flow between frames, uses it to guide feature propagation, and is the state of the art on RADIATE.
We use RaFD and SIRA as baselines and, since neither releases code, we reproduce them from scratch.

\subsection{Doppler Velocity from Spinning Radar}
\label{sec:related-doppler}

The Doppler shift distorts spinning radar scans, and \citet{burnett_we_2021} find that accounting for Doppler distortion improves localization. We show in this work that doing so improves detection as well.
Commercial spinning radars did not provide Doppler information until recent firmware changes~\cite{rennie_doppler_2023, lisus_are_2024}, which alternate up-chirps and down-chirps between azimuths, so that radial velocity appears as a range shift between neighbouring azimuths.
Since then, spinning radar Doppler has been shown to improve ego-motion estimation~\cite{rennie_doppler_2023, lisus_are_2024,legentil_dro_2025}.

\subsection{Doppler Velocity in Oriented Box Tracking}
\label{sec:related-tracking}

Radar trackers have long integrated Doppler velocity.
On automotive and 4D imaging radar, \citet{tan_tracking_2023} cluster the points of a vehicle, fit a box, and feed the velocity solved from the points' Doppler information to a linear Kalman filter as a measurement.
Others model the Doppler of each point in the measurement likelihood~\cite{scheel_tracking_2019}, or update a lidar-detected box with the radial velocities of the radar points inside it~\cite{xu_radar-informed_2026}.
Such velocities can also initialize a new track, on radar~\cite{kellner_instantaneous_2014,gosala_rls_2020} and on FMCW lidar~\cite{zeng_simpler_2025}.
However, all of these rely on a Doppler velocity attached to sparse points, whereas for spinning radar, the Doppler velocity is encoded in the range shift between neighbouring azimuths crossing each vehicle box in a dense intensity image.
The closest work, RadarMOT~\cite{xu_radar-informed_2026}, takes its boxes from a lidar detector, whereas our detections and velocities come from the same spinning radar. This removes the need for a second sensor, but makes tracking harder at the radar's slower frame rate.
Learned methods treat Doppler velocity as a network input~\cite{cheng_centerradarnet_2023} or predict a velocity outright, regressed by the detector~\cite{yin_center-based_2021} or seeded from a predicted direction~\cite{yataka_sira_2024}.
Our prior is measured rather than predicted, so it needs no velocity labels to train and, as we show, closely matches ground-truth velocities.

\section{Methodology}
\label{sec:method}

\subsection{Lidar-to-Radar Auto-Labelling}
\label{sec:method-labels}

Our labelling pipeline uses lidar observations to generate vehicle boxes, track identities, and velocities for spinning radar data.
It runs offline on each sequence of Boreas-RT~\cite{lisus_boreas_2026}, using its \SI{10}{\hertz}, 128-beam Velodyne Alpha Prime lidar, \SI{4}{\hertz} Navtech RAS6 spinning radar, calibrated sensor extrinsics, and post-processed GNSS--INS poses.
Figure~\ref{fig:labelling} follows the same vehicles through four stages: detect and fuse, track offline, fit trajectories, and transfer to radar.

Before detection, we compensate for ego-motion distortion by deskewing each lidar scan to its midpoint timestamp using the constant body-rate model of \citet{burnett_boreas_2023}. 

\subsubsection{Detect and fuse}
We adapt the MS3D++ ensemble framework to obtain vehicle proposals from multiple detectors and views.
Our ensemble contains an anchor-head and a center-head PV-RCNN++ detector~\cite{shi_pvrcnnpp_2023}.
Both are pretrained on Waymo~\cite{sun_scalability_2020}, whose lidar density is the closest to that of Boreas-RT and which therefore transfers better than nuScenes~\cite{caesar_nuscenes_2020} or Lyft~\cite{kesten_lyft_2019}.
We fine-tune these detectors for ten epochs on around \num{5000} annotated frames of the Boreas dataset~\cite{burnett_boreas_2023}, the only annotated in-domain data available.
At inference, each detector processes one, two, and four accumulated sweeps, both with and without test-time augmentation, yielding 12 detection views per frame. The dashed cyan boxes in Fig.~\ref{fig:labelling} illustrate the disagreement in position and orientation among these proposals.
Kernel-based fusion groups nearby box centres within a \SI{0.5}{\metre} radius, discards groups supported by fewer than three views, and estimates box position, dimensions, heading, and score from kernel density peaks.
MS3D++ then applies temporal filtering to remove poor detections and propagates confident static boxes through brief occlusions, finalizing the solid cyan boxes.
We retain detections with confidence scores above 0.3, which later stages re-score.

\subsubsection{Track offline}
We associate the fused boxes with MCTrack~\cite{wang_mctrack_2024} in offline mode.
Boxes are transformed to a common world frame, where a constant-velocity Kalman filter predicts each vehicle's planar motion and Hungarian matching associates detections.
The offline pass interpolates gaps across each track and assigns each track a single score shared by all its detections.
The output consists of discrete vehicle tracks with persistent identities and estimated velocities, shown by the coloured boxes, histories, and arrows in Fig.~\ref{fig:labelling}.

\subsubsection{Fit trajectories}
The discrete tracks still contain frame-to-frame variation in box position and heading.
For example, the orange track in Fig.~\ref{fig:labelling} wobbles from side to side.
We refine each track by fitting a continuous-time Gaussian-process trajectory with a white-noise-on-acceleration prior~\cite{anderson_full_2015} to the poses and body-frame velocities.
This enforces smooth, physically feasible motion across the entire track.
We also refine the track confidence score based on how closely MCTrack boxes align with the smoothed trajectory.
Only tracks with a final score of at least $0.8$ are transferred to radar to maximize label quality.
This discards tracks of high-confidence but poorly aligned boxes while keeping tracks of low-confidence boxes that form feasible trajectories.

\subsubsection{Transfer to radar}
A spinning radar acquires each scan with a separate timestamp for each azimuth.
Consequently, vehicles in different azimuths are observed at different times in the same image.
Transferring boxes directly, from the midpoint of the nearest lidar frame to the midpoint of a given radar frame, ignores both box motion and ego motion during the scan.
We account for both in our transfer algorithm.

For each radar scan and lidar track pair, we first find their temporal overlap and sample the fitted continuous-time vehicle trajectory at the overlap endpoints.
We then transform these positions into radar coordinates, look up the beam timestamps at their corresponding azimuths, and take the average as the sampling time $t^{\star}$.
The continuous trajectory provides the box pose and velocity at $t^{\star}$, and using sensor poses interpolated to this time, we transfer these quantities to radar coordinates as
\begin{equation}
  \mbf{T}_R(t^{\star}) =
  \mbf{T}_{WR}^{-1}(t^{\star})\mbf{T}_{WL}(t^{\star})
  \mbf{T}_L(t^{\star}),
  \label{eq:transfer-pose}
\end{equation}
\begin{equation}
  \mbf{v}_R(t^{\star}) =
  \mbf{C}_{WR}^{\trans}(t^{\star})\mbf{C}_{WL}(t^{\star})
  \mbf{v}_L(t^{\star}),
  \label{eq:transfer-velocity}
\end{equation}
where $W$, $L$, and $R$ denote world, lidar, and radar frames, $\mbf{T}\in\mathrm{SE}(3)$ is the box pose, and $\mbf{v}$ is its absolute velocity.
The interpolated sensor poses $\mbf{T}_{WL}$ and $\mbf{T}_{WR}$ map lidar and radar coordinates into the world frame, with rotations $\mbf{C}_{WL}$ and $\mbf{C}_{WR}$.
Evaluating the fitted vehicle trajectory and radar pose at $t^{\star}$ accounts for target and ego motion, respectively.

\subsection{Doppler Undistortion for Detection}
\label{sec:method-detection}

\begin{figure}[!t]
\centering
\begingroup
\def\undistortassetdir{tikz_images/doppler_undistort}%
\definecolor{undistortborder}{HTML}{E4E9EF}
\definecolor{undistortnearest}{HTML}{E69F00}
\definecolor{undistorttrajectory}{HTML}{0092A6}
\definecolor{undistortbracket}{HTML}{FFF3A0}
\begin{tikzpicture}[
  heading/.style={anchor=south west,inner sep=0pt,text=black,
    font=\fontsize{7}{8}\selectfont\bfseries},
]
  \foreach \x/\file/\heading in {0/rot_raw/(a) Raw,4.55/rot_gt/(b) Doppler-undistorted}{
    \node[anchor=south west,inner sep=0pt] at (\x,0)
      {\includegraphics[width=4.3cm]{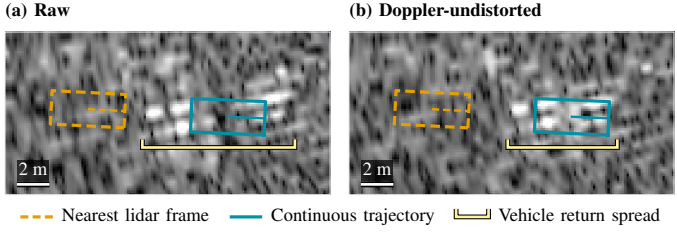}};
    \draw[draw=undistortborder,line width=.35pt] (\x,0) rectangle ++(4.3,2.15);
    \node[heading] at (\x,2.31) {\heading};
  }
  \node[anchor=north,inner sep=0pt,font=\fontsize{7}{8}\selectfont] at (4.425,-.15) {%
    \tikz[baseline=-.5ex]\draw[undistortnearest,line width=1.2pt,
      dash pattern=on 3pt off 1.6pt,line cap=butt] (0,0)--(12.2pt,0);%
    \hspace{1mm}Nearest lidar frame\hspace{2.5mm}%
    \tikz[baseline=-.5ex]\draw[undistorttrajectory,line width=1.2pt] (0,0)--(12.2pt,0);%
    \hspace{1mm}Continuous trajectory\hspace{2.5mm}%
    \tikz[baseline=-.5ex]{\draw[black,line width=1.8pt,line cap=butt,line join=miter] (0,3pt)--(0,0)--(12.2pt,0)--(12.2pt,3pt);
      \draw[undistortbracket,line width=1pt,line cap=butt,line join=miter] (0,3pt)--(0,0)--(12.2pt,0)--(12.2pt,3pt);}%
    \hspace{1mm}Vehicle return spread};
\end{tikzpicture}
\caption{\texttt{Skyway} scene of Boreas-RT with a vehicle travelling at \SI{32.1}{\metre/\second}, (a) raw and (b) after Doppler undistortion with the ground-truth ego velocity. Orange dashed boxes transfer the nearest lidar frame at its midpoint, and teal solid boxes sample the vehicle's continuous trajectory at the estimated azimuth-crossing time. The two boxes are separated by \SI{8.6}{\metre}. The brackets mark the spread of the vehicle's return. Note how the vehicle's return in the undistorted scan fit tighter within the teal box, with the remaining offset due to the target's own motion. Intensity is normalized per panel.}
\label{fig:undistort}
\endgroup
\end{figure}

Doppler distortion shifts radar returns along each azimuth, both from the ego vehicle's motion, which we correct, and from each target's own motion, which remains.
We apply the ego-Doppler correction of \citet{burnett_we_2021} using the Boreas devkit~\cite{lisus_boreas_2026}, with radar-derived ego velocity estimated following \citet{lisus_are_2024}.
This requires the alternating chirps of Boreas-RT's radar, which RADIATE's lacks, so RADIATE serves only to validate our detector implementations and all Doppler experiments use Boreas-RT.
We compare uncorrected scans with scans corrected using either estimated ego velocity or ground-truth GNSS--INS ego velocity.
Figure~\ref{fig:undistort} shows the raw shifts straddling the box and collapsing toward it after.

\begin{figure*}[!t]
  \centering
  \includegraphics[width=\textwidth]{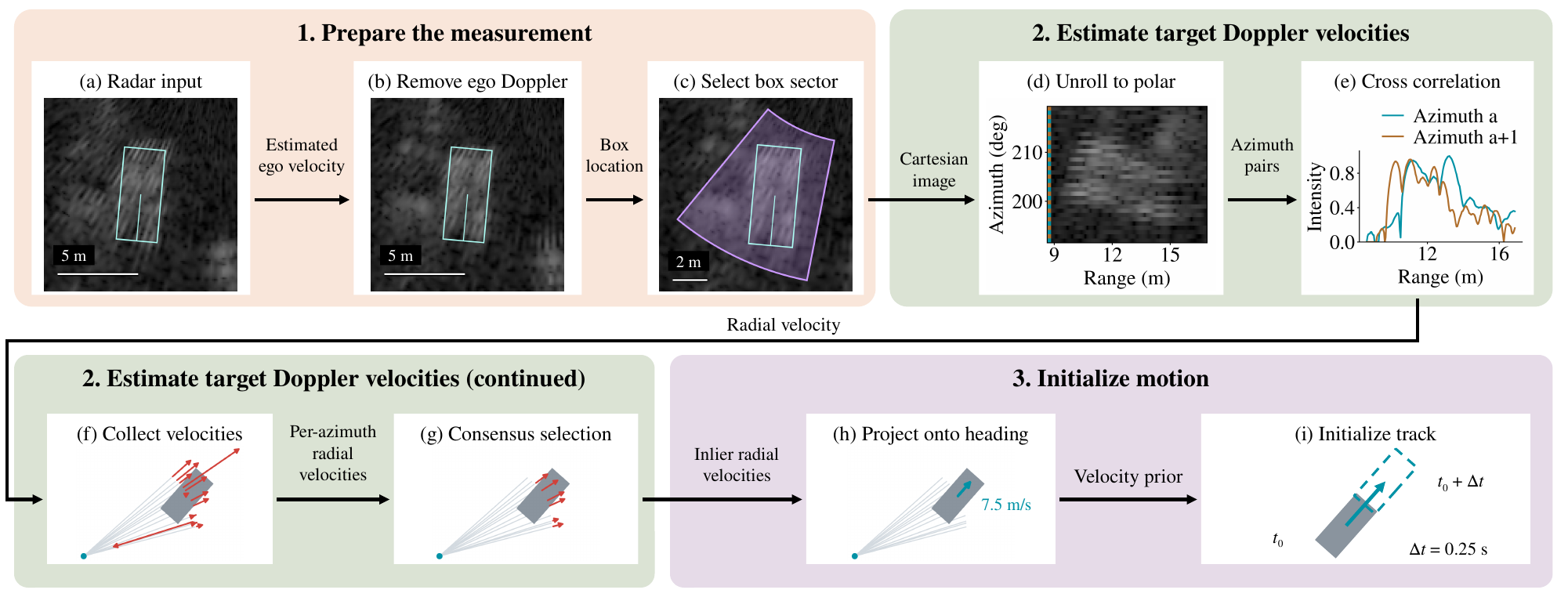}
\caption{Single-scan Doppler velocity initialization for tracking, following one vehicle through three stages. (a--c) Doppler undistortion using the estimated ego velocity and an annular sector (purple) around a detection box (cyan). (d--g) Cross-correlation within the polar rectangle and consensus selection of per-azimuth radial velocities (red). (h--i) Projection onto the box heading, giving the velocity prior (teal), and track initialization. This Boreas-RT vehicle is located \SI{12.8}{\metre} from the radar, with an estimated speed of \SI{7.5}{\metre/\second} against a ground-truth \SI{8.8}{\metre/\second}.}
  \label{fig:doppler-extraction}
\end{figure*}

\subsection{Doppler Velocity Prior for Tracking}
\label{sec:method-tracking}

A new radar track has no position history from which to estimate velocity.
At \SI{4}{\hertz}, a vehicle travelling at \SI{20}{\metre/\second} moves \SI{5}{\metre} between scans, so a zero-velocity initialization can place its first prediction far from the next detection.
We use the residual Doppler pattern after ego-motion correction to estimate a target vehicle's absolute velocity from a single scan and initialize MCTrack with this prior.
Figure~\ref{fig:doppler-extraction} organizes the procedure into three stages: prepare the measurement, estimate target Doppler velocities, and initialize motion.

\subsubsection{Prepare the measurement}
After the ego-Doppler correction of Section~\ref{sec:method-detection}, we draw an annular sector around the target, buffered by \SI{1}{\metre} in range and cross-range.
The buffer captures returns that the target's uncorrected Doppler shift displaces outside the box, while limiting surrounding clutter.

\subsubsection{Estimate target Doppler velocities}
Because the sector is a rectangle in polar coordinates, we apply the adjacent-azimuth cross-correlation of \citet{lisus_are_2024}, which Section~\ref{sec:method-detection} uses for ego velocity, within the sector.
Since clutter can produce competing correlation peaks, and far fewer azimuths cross a vehicle than are available for ego-velocity estimation, we keep up to three correlation maxima per pair. Panel (f) illustrates the resulting radial measurements across the sector.

As in panel (g), we select the velocity by a RANSAC-like~\cite{fischler_random_1981} consensus process, but evaluate every candidate rather than sampling hypotheses because the candidate set is small.
Each azimuth pair votes with its strongest correlation peak, and the candidate with the most votes within \SI{3}{\metre/\second} forms the inlier cluster.
To minimize noise, we require the winning cluster to contain at least half of the pairs; otherwise, no estimate is produced.
If this check passes, each pair then contributes its closest candidate within the same tolerance, so secondary peaks can recover inconsistent measurements.

\subsubsection{Initialize motion}
We make the assumption that motion is aligned with the longitudinal vehicle axis, and estimate the signed longitudinal speed $\hat v$ by least squares from the radial velocities $d_i$ (positive for approaching) at azimuths $\theta_i$, given the detected bounding box heading $\psi$, as shown in panel (h):
\begin{equation}
  \hat v=-\frac{\sum_{i\in\mathcal I}c_i d_i}
                              {\sum_{i\in\mathcal I}c_i^2},
  \label{eq:velocity-projection}
\end{equation}
where $c_i=\cos(\theta_i-\psi)$ and $\mathcal I$ is the set of all measurements satisfying $|c_i|\geq\cos(75^{\circ})$ to exclude near-broadside observations.
The velocity prior is then $\hat{\mbf{v}}=\hat v[\cos\psi,\sin\psi]^{\trans}$.
We apply a \SI{2}{\metre/\second} deadband, setting the prior to zero when $|\hat v|$ falls below this threshold or no valid estimate is available. 

We initialize each new track with the detected position and this velocity in world coordinates (panel (i)).

\section{Experiments}
\label{sec:experiments}

We present the labels from our pipeline, then use them to evaluate Doppler undistortion for detection and the estimated prior for tracking.

\subsection{Automatic Lidar-to-Radar Labelling}
\label{sec:exp-labels}

\subsubsection{Generated labels}
We summarize the generated labels here and assess them in the two parts that follow. The pipeline produced over \num{9.1} million lidar boxes and over \num{3.7} million radar boxes across 60 Boreas-RT sequences spanning \SI{643}{\kilo\metre} of driving.
Table~\ref{tab:dataset} reports the final outputs by route category.

\begin{table}[!t]
  \centering
  \caption{Final auto-generated labels by route. The grey routes are used to train the RaFD and SIRA detectors. Target speed is the absolute speed of surrounding vehicles, including stationary vehicles.}
  \label{tab:dataset}
  \scriptsize
  \renewcommand{\arraystretch}{1.15}
  \setlength{\tabcolsep}{1pt}
  \begin{tabularx}{\columnwidth}{l*{9}{>{\centering\arraybackslash}X}}
    \toprule
    & & \multicolumn{2}{c}{Frames (k)} & \multicolumn{2}{c}{Boxes (k)} & \multicolumn{2}{c}{Tracks (k)} & \multicolumn{2}{c}{Averages} \\
    \cmidrule(lr){3-4}\cmidrule(lr){5-6}\cmidrule(lr){7-8}\cmidrule(lr){9-10}
    Route & \makecell[c]{\# of\\seqs.} & Lidar & Radar & Lidar & Radar & Lidar & Radar & \makecell[c]{Seq.\\length\\(km)} & \makecell[c]{Target\\speed\\(m/s)} \\
    \midrule
    \rowcolor{black!7} \texttt{Regional} & 6 & 49.3 & 20.5 & 930.4 & 388.3 & 8.4 & 8.4 & 9.3 & 4.3 \\
    \rowcolor{black!7} \texttt{Skyway} & 5 & 28.8 & 12.0 & 370.6 & 153.9 & 5.3 & 5.2 & 11.1 & 9.7 \\
    \rowcolor{black!7} \texttt{Suburbs} & 10 & 103.3 & 43.1 & 2,195 & 913.2 & 18.0 & 18.0 & 7.9 & 2.1 \\
    \rowcolor{black!7} \texttt{Urban} & 7 & 164.5 & 68.4 & 2,886 & 1,203 & 14.2 & 14.2 & 8.6 & 2.4 \\
    \texttt{Industrial} & 5 & 43.4 & 18.1 & 803.7 & 335.9 & 5.5 & 5.5 & 5.4 & 1.8 \\
    \texttt{Tunnel} & 10 & 20.3 & 8.5 & 170.1 & 70.6 & 1.4 & 1.4 & 1.9 & 6.7 \\
    \texttt{Farm} & 10 & 101.2 & 42.1 & 75.0 & 31.1 & 0.8 & 0.8 & 10.8 & 5.5 \\
    \texttt{Forest} & 4 & 43.6 & 18.1 & 23.5 & 9.6 & 0.4 & 0.4 & 16.4 & 3.6 \\
    \texttt{Freeway} & 3 & 97.3 & 40.5 & 1,666 & 692.8 & 18.7 & 18.7 & 57.6 & 9.0 \\
    \midrule
    Total & 60 & 651.8 & 271.3 & 9,120 & 3,798 & 72.6 & 72.5 & 10.7 & 4.1 \\
    \bottomrule
  \end{tabularx}
\end{table}

\subsubsection{Trajectory refinement}
\begin{figure}[!t]
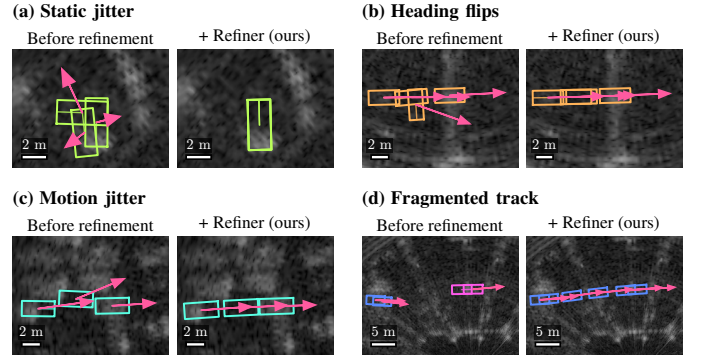

\centering
\begingroup
\def\historyassetdir{tikz_images/cttrack}%
\resizebox{\columnwidth}{!}{%
\begin{tikzpicture}[
  panel/.style={anchor=north west,inner sep=0pt},
  title/.style={anchor=south west,inner sep=0pt,font=\fontsize{7}{8.5}\selectfont\bfseries},
  method/.style={anchor=south,inner sep=0pt,font=\fontsize{6.5}{7.5}\selectfont},
]
  \foreach \case/\x/\y/\title in {a/0/0/Static jitter,b/4.57/0/Heading flips,c/0/-2.45/Motion jitter,d/4.57/-2.45/Fragmented track}{
    \node[title] at (\x,\y+.36) {(\case)\hspace{.3em}\title};
    \node[method] at (\x+1.02,\y+.05) {Before refinement};
    \node[method] at (\x+3.17,\y+.05) {+ Refiner (ours)};
    \node[panel] at (\x,\y) {\includegraphics[width=2.04cm]{\historyassetdir/\case_mctrack.pdf}};
    \node[panel] at (\x+2.15,\y) {\includegraphics[width=2.04cm]{\historyassetdir/\case_cttrack.pdf}};
  }
\end{tikzpicture}%
}%
\endgroup
  \caption{Four vehicle tracks from the \texttt{Urban} and \texttt{Regional} routes of Boreas-RT, before refinement (left) and with our refiner (right). Boxes are per-frame poses, coloured by track identity, and magenta arrows are the estimated velocity, suppressed below \SI{0.5}{\metre/\second}. The four failure modes are (a) static jitter, (b) heading flips, (c) motion jitter, and (d) fragmentation.}
  \label{fig:cttrack-refinement}
\end{figure}

Figure~\ref{fig:cttrack-refinement} shows four MCTrack failure modes and how refinement resolves them.
Refinement reduces the stationary car's planar root-mean-square spread about its mean centre from \SI{0.43}{\metre} to \SI{0.04}{\metre} (a), resolves a $180^{\circ}$ heading flip that MCTrack interpolated into a physically impossible sideways car (b), removes a \SI{1.2}{\metre} lateral jump caused by a poor detection (c), and merges two fragments of the same car separated by \SI{2.39}{\second} (d).
MCTrack corrects each frame on its own, whereas the trajectory fit constrains the whole track.
A heading flipped for an entire track, or a vehicle the detector never sees, remains uncorrected, since refinement can only smooth the evidence the detector provides. 
To quantify the refinement, we evaluate tracking before and after our refiner on 749 human-labelled Boreas frames held out from detector fine-tuning. Refinement raises MOTA from 73.1 to 76.7, cuts fragmentations from 147 to 90, and reduces misses from \num{2255} to \num{1481}.

\subsubsection{Transfer to radar}
Figure~\ref{fig:undistort} compares nearest-frame transfer with sampling the continuous trajectory at its estimated radar azimuth-crossing time.
Both overlays originate from the same refined lidar track, so the comparison isolates the transfer step.
The vehicle travels at \SI{32.1}{\metre/\second} and the ego vehicle at \SI{27.3}{\metre/\second}, and the two transfers differ in box centre by \SI{8.6}{\metre}, which combines the target's motion over the azimuth-time offset with the ego motion over that interval.
The sampled box covers the visible radar returns, whereas the nearest-frame box is displaced along the direction of travel, which is why nearest-frame transfer fails for fast traffic.

\subsection{Doppler Undistortion for Detection}
\label{sec:exp-detection}

\subsubsection{Detector reproduction}
We first assess our four-frame RaFD and SIRA implementations on RADIATE to verify their correctness, using the good-weather and good+bad-weather splits prescribed by its authors~\cite{sheeny_radiate_2020}.
Our evaluator computes single-class vehicle mean average precision (mAP) in bird's-eye view at intersection-over-union (IoU) thresholds $0.3$, $0.5$, and $0.7$, denoted $\map@\tau$, with a confidence threshold of $0.05$.
Table~\ref{tab:radiate} reports the results next to the published scores.

\begin{table}[!t]
  \centering
  \caption{RaFD and SIRA reproduction on RADIATE. We report mean $\pm$ sample standard deviation over three seeds. Bold marks the better of the published~\cite{yang_rafd_2025,yataka_sira_2024} and reproduced scores.}
  \label{tab:radiate}
  \footnotesize
  \setlength{\tabcolsep}{2.2pt}
  \medmuskip=1mu
  \resizebox{\columnwidth}{!}{%
  \begin{tabular}{@{}lccc@{\hspace{6pt}}ccc@{}}
    \toprule
    & \multicolumn{3}{c}{Good weather} & \multicolumn{3}{c}{Good + bad weather} \\
    \cmidrule(lr){2-4}\cmidrule(lr){5-7}
    Method & $\map@0.3$ & $\map@0.5$ & $\map@0.7$ & $\map@0.3$ & $\map@0.5$ & $\map@0.7$ \\
    \midrule
    RaFD & $\mathbf{69.4\pm1.5}$ & $\mathbf{59.5\pm1.9}$ & $\mathbf{23.9\pm1.1}$ & $68.8\pm1.8$ & $54.2\pm1.7$ & $18.8\pm1.6$ \\
    Our RaFD & $67.8\pm2.0$ & $56.9\pm1.5$ & $20.0\pm1.6$ & $\mathbf{72.0\pm2.8}$ & $\mathbf{59.7\pm2.6}$ & $\mathbf{20.3\pm0.9}$ \\
    \midrule
    SIRA & $68.7\pm1.1$ & $58.1\pm1.4$ & $22.8\pm0.9$ & $66.1\pm0.8$ & $53.8\pm1.1$ & $19.9\pm1.0$ \\
    Our SIRA & $\mathbf{69.4\pm1.8}$ & $\mathbf{58.7\pm2.3}$ & $\mathbf{23.9\pm1.7}$ & $\mathbf{68.6\pm0.6}$ & $\mathbf{55.0\pm0.2}$ & $\mathbf{20.4\pm0.5}$ \\
    \bottomrule
  \end{tabular}%
  }
\end{table}

Although our implementations do not exactly reproduce the published results, they reach comparable performance, which supports their use in the following experiments.

\subsubsection{Effect of Doppler undistortion on detection}
The 28 grey Boreas-RT sequences in Table~\ref{tab:dataset} are split into 20 for training, 4 for validation, and 4 for testing, with one complete validation and test sequence held out per route.
We train RaFD and SIRA on these data under three radar preprocessing conditions: raw images, Doppler undistortion using ground-truth (GT) ego velocity, and Doppler undistortion using radar-estimated ego velocity.
We evaluate each condition on the four Boreas-RT test sequences, containing \num{22752} radar scans and \num{458376} radar boxes.
Within each detector, the four-frame inputs, labels, training recipe, and seeds are matched across conditions.
We use Adam with learning rate $2\times10^{-4}$, batch size 4, and a ten-epoch cosine schedule.
Table~\ref{tab:egodopp} reports the results.

\begin{table}[!t]
  \centering
  \caption{Detection results on Boreas-RT. We report mean $\pm$ sample standard deviation over three seeds. GT and estimated refer to the ego velocity used for Doppler undistortion. Bold marks the best input per detector.}
  \label{tab:egodopp}
  \footnotesize
  \setlength{\tabcolsep}{3pt}
  \begin{tabular}{lccc}
    \toprule
    Input & $\map@0.3$ & $\map@0.5$ & $\map@0.7$ \\
    \midrule
    RaFD, raw & $83.82\pm0.10$ & $80.21\pm0.10$ & $65.13\pm0.20$ \\
    \quad + Estimated ego & $\mathbf{84.85\pm0.05}$ & $81.45\pm0.13$ & $67.50\pm0.10$ \\
    \quad + GT ego & $84.75\pm0.13$ & $\mathbf{81.50\pm0.20}$ & $\mathbf{67.65\pm0.17}$ \\
    \midrule
    SIRA, raw & $76.46\pm0.35$ & $73.00\pm0.35$ & $59.44\pm0.56$ \\
    \quad + Estimated ego & $76.28\pm0.86$ & $73.15\pm0.88$ & $60.44\pm1.21$ \\
    \quad + GT ego & $\mathbf{76.57\pm0.38}$ & $\mathbf{73.40\pm0.35}$ & $\mathbf{60.70\pm0.18}$ \\
    \bottomrule
  \end{tabular}
\end{table}

Doppler undistortion with radar-estimated ego velocity consistently improves RaFD, the stronger of the two detectors on Boreas-RT, with improvements in every seed at every threshold.
The gain grows with the overlap threshold, reaching $2.37$ mAP points at IoU $0.7$ against $2.52$ with GT ego velocity, which is consistent with better localization after removing the ego-dependent range shift.
SIRA responds less consistently.
Its estimated-ego changes lie within its seed standard deviation at every threshold, while GT-ego undistortion improves it by $1.26$ points at IoU $0.7$ with little seed variation.
Across both detectors, estimated-ego and GT-ego means differ by at most $0.29$ mAP points, so the ego-velocity estimate is not the limiting factor.
We hypothesize that RaFD's stronger localization allows it to exploit the more consistent radar appearance after undistortion.
Overall, the results show a modest detection benefit that is consistent for RaFD and within seed noise for SIRA.

\subsection{Doppler Velocity Prior for Tracking}
\label{sec:exp-tracking}

\subsubsection{Accuracy of the velocity prior}
\begin{figure}[!t]
  \centering
  \includegraphics[width=\columnwidth]{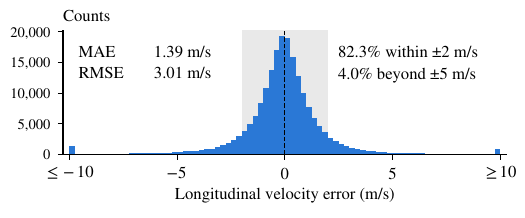}
  \caption{Longitudinal velocity errors on \num{199253} estimated vehicles from \texttt{Skyway}, \texttt{Suburbs}, and \texttt{Urban} sequences. The shaded band marks errors within \SI{2}{\metre/\second}, and errors outside $[-10,10]$~\si{\metre/\second} accumulate in the boundary bins. Mean absolute error (MAE) and root-mean-square error (RMSE) use the untruncated errors and differ from the planar $\maev$ used for tracking.}
  \label{fig:vehicle-accuracy}
\end{figure}

We first verify that the single-scan per-box velocity estimate is accurate enough to serve as a prior. Figure~\ref{fig:vehicle-accuracy} aggregates the scalar error between the longitudinal velocity produced by our method and the ground-truth velocity projected onto the longitudinal axis.
The estimator returns a value for \num{199253} of the \num{243999} vehicle observations ($81.7\%$) in all \texttt{Skyway}, \texttt{Suburbs}, and \texttt{Urban} sequences. The rest are rejected for insufficient Doppler consensus or near-broadside viewing.

The error is unimodal and centred near zero, with $82.3\%$ of estimates within \SI{2}{\metre/\second} of the ground truth.
Despite the heavy tails, the estimate is still a useful prior.

\subsubsection{Tracking with ground-truth boxes}
We evaluate seven test sequences with \num{852000} radar ground-truth boxes, adding \texttt{Freeway}, \texttt{Tunnel}, and \texttt{Industrial} routes to the four detector test sequences.
We supply these boxes directly to MCTrack and compare zero init., our Doppler prior, and a privileged GT-velocity prior.
We report MOTA~\cite{bernardin_evaluating_2008}, which measures tracking accuracy, and the identity F1 score (IDF1)~\cite{ristani_performance_2016}, which measures identity precision and recall.
We also report identity switches (IDSW) per \num{1000} reference boxes and $\maev$, the mean Euclidean error between matched predicted and reference planar velocities.

\begin{figure}[!t]
  \centering
  \includegraphics[width=\columnwidth]{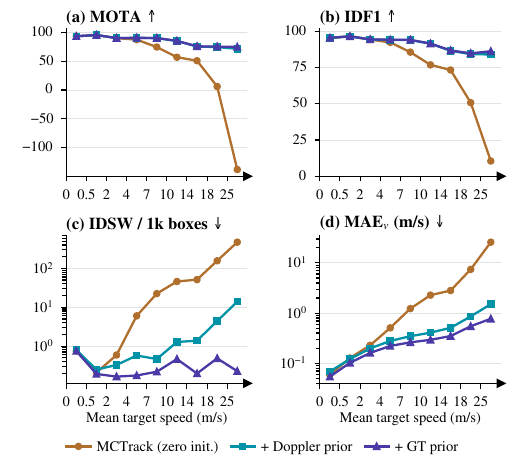}
  \caption{Tracking with GT boxes on seven Boreas-RT test sequences, varying only the birth velocity. Frames are grouped by mean reference-vehicle speed. MOTA is multi-object tracking accuracy, IDF1 is the identity F1 score, IDSW is identity switches per \num{1000} reference boxes, and $\maev$ is the mean planar velocity error. Zero init. diverges from both priors above \SI{10}{\metre/\second}.}
  \label{fig:tracking-gt-bins}
\end{figure}

Figure~\ref{fig:tracking-gt-bins} groups frames by mean reference-vehicle speed.
At low speeds all three behave similarly, but at high speeds, zero init. produces sharply more identity switches and larger velocity errors, while the Doppler prior follows the GT prior.
In the highest-speed bin ($\geq\SI{25}{\metre/\second}$), zero init. collapses to a MOTA of $-138.1$ and a velocity error of \SI{25.96}{\metre/\second}, while the Doppler prior remains at $71.6$ and \SI{1.51}{\metre/\second}, against $74.8$ for the GT prior.
Negative MOTA means false positives, misses, and identity switches exceed the reference-box count.
The failure begins near \SI{10}{\metre/\second}, where the \SI{2.5}{\metre} inter-scan displacement approaches the size of a vehicle box and association by overlap starts to fail. The Doppler prior tracks the GT prior because its typical velocity error of \SIrange{1}{2}{\metre/\second} (Fig.~\ref{fig:vehicle-accuracy}) shifts the predicted position by well under a metre.

\subsubsection{Tracking with detector predictions}
To test whether the gain survives detector errors, we also track RaFD and SIRA predicted boxes.
The detectors run on images with radar-estimated ego-Doppler correction.

\begin{figure}[!t]
  \centering
  \includegraphics[width=\columnwidth]{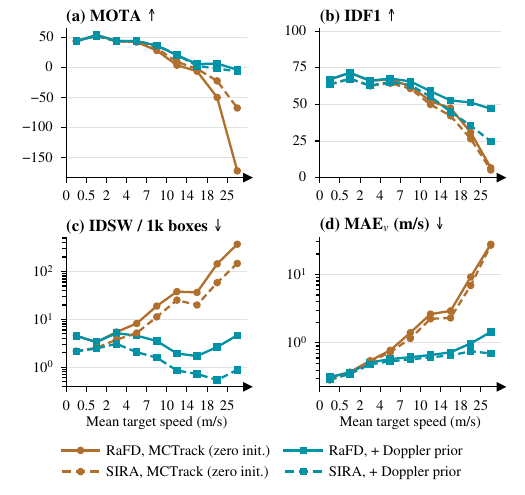}
  \caption{Tracking RaFD~\cite{yang_rafd_2025} (solid) and SIRA~\cite{yataka_sira_2024} (dashed) detections on the same seven test sequences as Fig.~\ref{fig:tracking-gt-bins}, with zero init. (circles) or the Doppler prior (squares). Speed bins and metrics match Fig.~\ref{fig:tracking-gt-bins}. No GT-velocity prior is available for detector boxes.}
  \label{fig:frame-bins}
\end{figure}

Figure~\ref{fig:frame-bins} shows that the benefit persists with both detectors, particularly in faster traffic.
For frames averaging \SIrange{18}{25}{\metre/\second}, the prior raises RaFD's MOTA from $-49.3$ to $7.0$ and SIRA's from $-21.8$ to $-1.6$, and in the highest-speed bin the velocity errors fall from about \SI{27}{\metre/\second} to below \SI{1.5}{\metre/\second} for both.
Unlike with ground-truth boxes, missed detections and false positives limit the resulting tracks, so the fastest bin's MOTA remains slightly negative even with the prior.
The prior therefore improves motion initialization and data association even with imperfect detections.

\begin{table}[!t]
  \centering
  \caption{Tracking on seven test sequences with \num{852000} ground-truth reference boxes. Bold marks the best prior per source. HOTA, DetA, and AssA are higher order tracking accuracy and its detection and association components, MOTA multi-object tracking accuracy, IDF1 identity F1, IDSW identity switches, Frag. fragmentations, FP/FN false positives/negatives, and $\maev$ the mean planar velocity error in \si{\metre/\second}.}
  \label{tab:tracking}
  \fontsize{6.5}{7.5}\selectfont
  \setlength{\tabcolsep}{0pt}
  \begin{tabular*}{\columnwidth}{@{\extracolsep{\fill}}lrrrrrrrrrc@{}}
    \toprule
    Prior & HOTA$\uparrow$ & DetA$\uparrow$ & AssA$\uparrow$ & MOTA$\uparrow$ & IDF1$\uparrow$ & IDSW$\downarrow$ & Frag.$\downarrow$ & FP$\downarrow$ & FN$\downarrow$ & $\maev\downarrow$ \\
    \midrule
    \multicolumn{11}{@{}l}{\textit{Ground-truth boxes}} \\
    Zero init.  & 85.16 & 80.66 & 90.06 & 76.86 & 86.41 & 25,511 & 6,478 & 117,232 & 54,440 & 1.31 \\
    Doppler     & 91.39 & 89.40 & 93.48 & 90.54 & 94.12 & 1,001 & 2,424 & 40,235 & 39,384 & 0.25 \\
    GT vel.     & \textbf{91.57} & \textbf{89.57} & \textbf{93.68} & \textbf{90.79} & \textbf{94.29} & \textbf{232} & \textbf{2,336} & \textbf{39,441} & \textbf{38,837} & \textbf{0.18} \\
    \midrule
    \multicolumn{11}{@{}l}{\textit{RaFD detections}} \\
    Zero init.  & 51.26 & 45.23 & 58.71 & 32.21 & 61.51 & 23,298 & 52,552 & 301,007 & 253,293 & 1.64 \\
    Doppler     & \textbf{53.75} & \textbf{48.23} & \textbf{60.49} & \textbf{42.61} & \textbf{66.35} & \textbf{3,534} & \textbf{52,437} & \textbf{232,930} & \textbf{252,524} & \textbf{0.50} \\
    \midrule
    \multicolumn{11}{@{}l}{\textit{SIRA detections}} \\
    Zero init.  & 49.63 & 42.25 & 58.71 & 37.63 & 59.83 & 10,686 & 40,598 & 162,863 & 357,849 & 1.07 \\
    Doppler     & \textbf{50.78} & \textbf{43.52} & \textbf{59.66} & \textbf{41.86} & \textbf{62.06} & \textbf{1,966} & \textbf{40,207} & \textbf{136,103} & \textbf{357,258} & \textbf{0.44} \\
    \bottomrule
  \end{tabular*}
\end{table}

Table~\ref{tab:tracking} aggregates all seven sequences and adds higher order tracking accuracy (HOTA), its detection (DetA) and association (AssA) components~\cite{luiten_hota_2021}, and error counts.
With ground-truth boxes, the Doppler prior improves MOTA by $13.68$ points and reduces identity switches by $96.1\%$, recovering $99.7\%$ of the GT prior's MOTA.
With detector boxes, the prior improves every reported metric for both detectors.
Since the detector outputs are identical, RaFD's \num{68077} fewer false positives come entirely from better association.

\subsubsection{Qualitative comparison}
Figure~\ref{fig:qual-tracking} shows the effect.
With zero init., a fast vehicle's displacement between scans defeats association, so a new track is born each frame while the old one persists, whereas the Doppler prior places the prediction near the next detection until the filter has enough history to estimate velocity on its own.
\section{Conclusion}
\label{sec:conclusion}
This paper addresses three problems in spinning radar vehicle detection and tracking, namely the lack of labelled training data, Doppler distortion, and large inter-frame motion at the radar's low scan rate.
Our automatic lidar-to-radar labelling pipeline provides novel temporally consistent labels for all \SI{643}{\kilo\metre} of the Boreas-RT dataset.
Using these labels, we find that ego-Doppler undistortion yields modest detection improvements by reducing range shifts.
For tracking, our single-scan vehicle-velocity extraction method provides motion information at track birth, substantially improving association with both ground-truth boxes and detector predictions.

The main limitation of our labelling pipeline is its reliance on lidar observations.
Vehicles visible to radar may be absent from the labels when occlusion, adverse weather, or airborne dust limits lidar visibility.
These missing labels can penalize valid radar detections during training and evaluation, limiting what a radar detector can learn beyond lidar.

Closing this gap is the next step, and future work will supplement the automatic labels with manual radar annotations in scenes where lidar visibility is limited.
Incorporating the uncertainty of the Doppler velocity estimate into the tracker, as a measurement rather than only a birth prior, would limit the influence of unreliable estimates and make fuller use of Doppler information.

\renewcommand*{\bibfont}{\footnotesize}

\printbibliography

\end{document}